\documentclass[letterpaper, 10 pt, journal, twoside]{IEEEtran}
\IEEEoverridecommandlockouts

\usepackage{cite}
\usepackage{amsmath,amssymb,amsfonts}
\usepackage{tikz}
\usetikzlibrary{shapes.geometric, arrows, positioning, fit, backgrounds}
\usepackage{makecell}
\usepackage{listings}
\usepackage{graphicx}\usepackage{textcomp}
\usepackage{xcolor}
\usepackage{makecell}
\usepackage{multirow, multicol} 
\usepackage{booktabs}

\def\BibTeX{{\rm B\kern-.05em{\sc i\kern-.025em b}\kern-.08em
    T\kern-.1667em\lower.7ex\hbox{E}\kern-.125emX}}

\title{\LARGE \bf
Through the Lens of Doubt: Robust and Efficient Uncertainty Estimation for Visual Place Recognition
}
\author{Emily Miller$^{1}$, Michael Milford$^{2}$, Muhammad Burhan Hafez$^{1}$, SD Ramchurn$^{1}$ and Shoaib Ehsan$^{1, 3}$%
\thanks{}%
\thanks{$^{1}$E. Miller, M. Burhan Hafez, S. D. Ramchurn and S. Ehsan are with the School of Electronics and Computer Science, University of Southampton, United Kingdom {\tt\small (email: em3g20@soton.ac.uk; burhan.hafez@soton.ac.uk; sdr1@soton.ac.uk, s.ehsan@soton.ac.uk)}}%
\thanks{$^{2}$M. Milford is with the School of Electronics and Computer Science, Queensland University of Technology, Brisbane, QLD 4000, Australia
        {\tt\small (email: michael.milford@qut.edu.au)}}%
\thanks{$^{3}$S. Ehsan is also with the School of Computer Science and Electronic Engineering, University of Essex, United Kingdom \tt\small{(email: sehsan@essex.ac.uk)}}%
}
\begin{document}

\maketitle

\begin{abstract}
Visual Place Recognition (VPR) enables robots and autonomous vehicles to identify previously visited locations by matching current observations against a database of known places. However, VPR systems face significant challenges when deployed across varying visual environments, lighting conditions, seasonal changes, and viewpoints changes. Failure-critical VPR applications, such as loop closure detection in simultaneous localization and mapping (SLAM) pipelines, require robust estimation of place matching uncertainty. We propose three training-free uncertainty metrics that estimate prediction confidence by analyzing inherent statistical patterns in similarity scores from any existing VPR method. Similarity Distribution (SD) quantifies match distinctiveness by measuring score separation between candidates; Ratio Spread (RS) evaluates competitive ambiguity among top-scoring locations; and Statistical Uncertainty (SU) is a combination of SD and RS that provides a unified metric that generalizes across datasets and VPR methods without requiring validation data to select the optimal metric. All three metrics operate without additional model training, architectural modifications, or computationally expensive geometric verification. Comprehensive evaluation across nine state-of-the-art VPR methods and six benchmark datasets confirms that our metrics excel at discriminating between correct and incorrect VPR matches,  and consistently outperform existing approaches while maintaining negligible computational overhead, making it deployable for real-time robotic applications across varied environmental conditions with improved precision-recall performance.

\end{abstract}


\section{Introduction}

\begin{figure}[h]
    \centering
    \includegraphics[width=\linewidth]{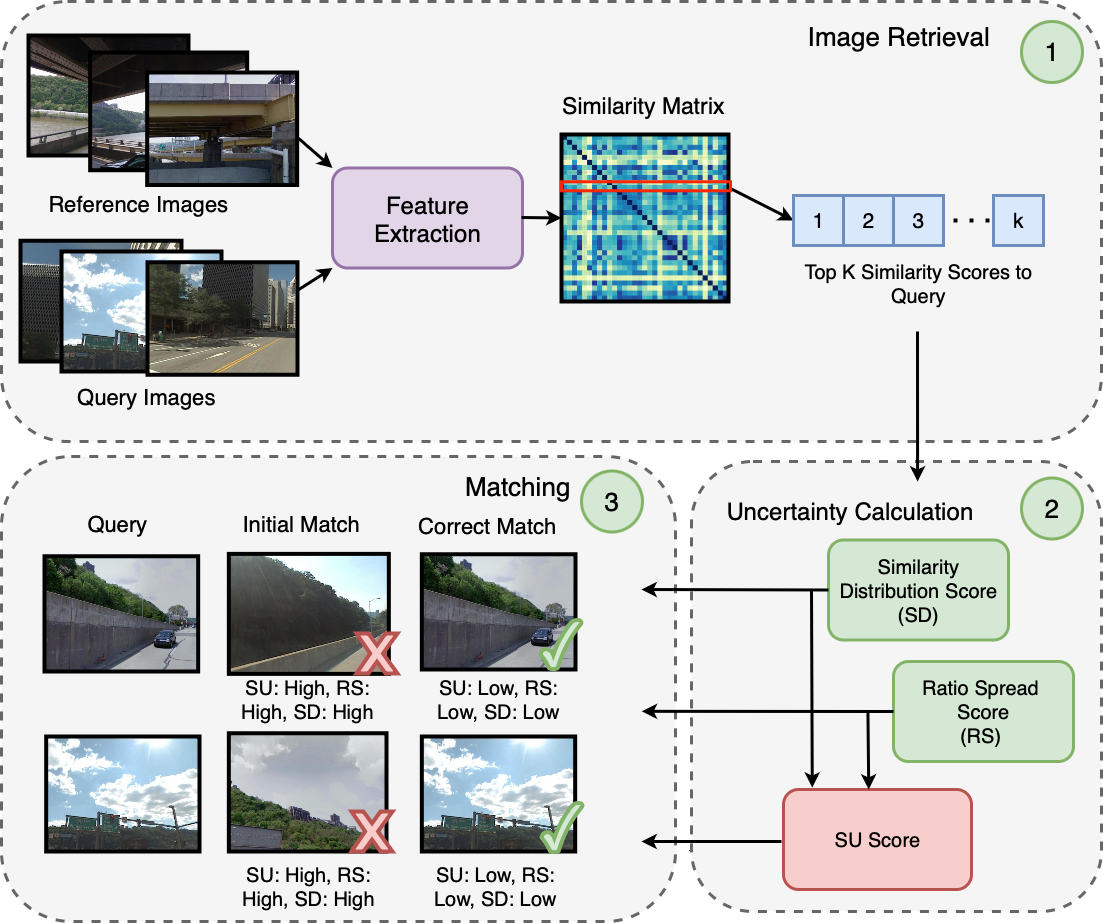}
    \caption{Overview of our training-free uncertainty estimation metrics integrated into the VPR pipeline, operating directly on similarity scores to provide reliable confidence estimates without requiring model modifications. Shown in Stage 3 in the diagram, examples of the initial match being wrong and having high uncertainty and the following image, which is correct, is the least uncertain image from the corresponding top-$k$ on the Pittsburgh250k dataset using Eigenplaces.}
    \label{fig:block-diagram}
    \vspace{-5mm}
\end{figure}

Visual Place Recognition (VPR) serves as a fundamental component in robotic localization and autonomous navigation systems \cite{SLAM}, enabling robots to determine their position by matching current observations with a database of georeferenced images \cite{survey}. Despite significant advances in descriptor-based methods \cite{tutorial}, real-world VPR deployment in failure-critical applications like autonomous vehicles, remains challenging due to inadequate uncertainty quantification. Mission-critical systems such as loop closure detection in SLAM pipelines \cite{whats_your_place} require robust confidence estimates to prevent catastrophic failures when VPR systems encounter perceptual aliasing, seasonal variations, and repetitive structures across different environmental conditions and camera viewpoints.

Current uncertainty estimation methods face practical limitations that prevent their use in failure-intolerant applications. Retrieval-based approaches such as L2 distance \cite{l2} and Perceptual Aliasing Score \cite{pa} show inconsistent performance across datasets and environments, with metric selection depending on the deployment context. Deep Uncertainty Estimation methods like Bayesian Triplet Loss \cite{blt} and STUN \cite{stun} require costly retraining for each new VPR architecture \cite{whats_your_place}, while Geometric Verification methods using SIFT \cite{sift}, DELF \cite{delf}, and SuperPoint \cite{superpoint} are computationally prohibitive for real-time operation. Selecting the most reliable metric typically requires validation data that may be unavailable in unseen environments, posing risks in safety-critical systems where incorrect high-confidence matches can lead to catastrophic failures \cite{general_survey}.


This work introduces training-free uncertainty estimation metrics that operate directly on VPR similarity scores without requiring retraining or architectural modifications. These metrics are derived from distributional analysis of top-$k$ matches: Similarity Distribution (SD) measures match distinctiveness by quantifying how clearly the top candidate separates from competing matches through the gap between highest and median similarity scores, while Ratio Spread (RS) evaluates competitive ambiguity by analyzing score clustering among top candidates. We further develop another metric, Statistical Uncertainty (SU) score, which is a combination of SD and RS, that eliminates the need for validation data to determine which individual metric performs best for specific VPR methods or datasets. These metrics maintain robust performance across diverse deployment scenarios while preserving the computational efficiency required for real-time operation.

Our main contributions are:
\begin{itemize}
    \item Three training-free uncertainty estimation metrics applicable to any VPR method: Similarity Distribution, Ratio Spread, and their combination (SU score) provides a unified metric that generalizes across VPR architectures and datasets without requiring validation data.
    \item We conduct a comprehensive two-stage evaluation encompassing nine VPR methods across six benchmark datasets. In the first stage, we assess the predictive capability of the proposed uncertainty metrics using AUC-PR, demonstrating their superior ability to identify matching failures. In the second stage, we evaluate their practical impact on end-to-end VPR performance by showing they measurably improve end VPR system performance, boosting precision at any given recall level when used as a rejection mechanism.
\end{itemize}

The rest of the paper is organized as follows. Section II reviews related work in VPR and uncertainty estimation. Section III introduces three uncertainty estimation metrics. Section IV outlines experimental setup. Results and analysis are given in Section V. Ablation studies are provided in Section VI. Finally, Section VII draws conclusions.

\section{Related Work}

\subsection{Visual Place Recognition}
Visual Place Recognition (VPR) has evolved from handcrafted local features like SIFT \cite{sift} and probabilistic frameworks such as FAB-MAP \cite{fabmap} to deep learning approaches that achieve superior robustness through automatically learned hierarchical representations \cite{tutorial}. Modern VPR architectures employ feature aggregation layers that compress spatial feature maps into compact global descriptors. NetVLAD \cite{netvlad} introduced differentiable aggregation using learnable cluster centers, while subsequent methods like MixVPR \cite{mixvpr} incorporate multi-scale feature mixing and TransVPR \cite{tansvpr} uses Transformer-based self-attention mechanisms. These architectures are typically trained using Siamese or Triplet networks with metric learning objectives \cite{siamese, Revaud2019}, enabled by large-scale datasets like GSV-Cities \cite{gsv-cities}.

\subsection{Uncertainty Estimation in Visual Place Recognition}
Robust uncertainty estimation proves crucial for safety-critical applications, where high-confidence incorrect matches from VPR systems can lead to catastrophic failures such as corrupted maps in Simultaneous Localization and Mapping (SLAM) pipelines \cite{SLAM, won2025}. Several works have demonstrated that rejecting predictions based on confidence estimates can significantly improve accuracy and prevent such failures \cite{Agarwal2025, SUE}, motivating the development of specialized uncertainty quantification methods for VPR systems. These VPR-specific methods draw inspiration from the broader deep learning community, where uncertainty is typically categorized as aleatoric (inherent data noise) and epistemic (model ignorance due to limited training data) \cite{kendall2017}. General techniques to capture this include Monte Carlo Dropout \cite{mc-dropout}, which approximates Bayesian inference by running multiple forward passes with active dropout layers, and Deep Ensembles \cite{ensembles}, which train multiple models to measure prediction variance. Other approaches modify the model's output to directly predict a probability distribution instead of a point estimate \cite{malinin2018}. However, these methods often impose significant computational overhead or require architectural modifications and extensive retraining, making them impractical for seamless integration with pre-existing, real-time VPR pipelines \cite{hendrycks2018}. Our work bypasses these limitations by operating directly on the output similarity scores of any VPR model.

VPR-specific uncertainty estimation methods address these challenges through four main categories. Retrieval-based Uncertainty Estimation (RUE) methods estimate uncertainty directly from similarity scores, such as L2 distance or Perceptual Aliasing (PA) score, which is computed as the ratio between first and second nearest neighbor distances \cite{SUE, l2, pa}. While computationally efficient and training-free, RUE methods suffer from overconfidence in perceptual aliasing scenarios \cite{SUE}.

Data-driven Uncertainty Estimation (DUE) methods learn to infer uncertainty from image content, typically modeling aleatoric uncertainty. Approaches like Bayesian Triplet Loss \cite{blt} and self-supervised methods such as STUN \cite{stun} exemplify this category but require specialized training and exhibit sensitivity to domain shifts across different environments.

Geometric Verification (GV) approaches assess spatial consistency between query and database images through local feature matching \cite{sift}. Methods employing local features like DELF \cite{delf} or SuperPoint \cite{superpoint} with RANSAC-based verification achieve high robustness but incur computational costs prohibitive for real-time applications. 

Spatial Uncertainty Estimation (SUE) methods \cite{SUE, pion2020} infer uncertainty from the spatial distribution of retrieved poses. Though effective in dense mapping scenarios, SUE performance depends on database quality and fails to detect visual ambiguity where geographically distinct locations appear similar. Additionally, SUE requires spatial pose information that may be unavailable at runtime.

Our metrics address these limitations by providing training-free uncertainty estimation that operates directly on similarity scores without model modifications, additional training, or external spatial information, while maintaining computational efficiency for real-time deployment.

Beyond VPR, similar principles for uncertainty estimation have been explored in object recognition \cite{object_det}, semantic segmentation \cite{sem_seg}, and aerial navigation \cite{aerial_nav}, where predictive uncertainty guides decision confidence and safety assessment. Our metrics extend these ideas to the retrieval-based VPR setting, emphasizing computational efficiency and training-free deployment.

\section{Proposed Uncertainty Estimation Metrics}
Uncertainty in VPR arises from two primary sources, each evident in the distribution of similarity scores. The first is perceptual aliasing, where visually similar but geographically distinct locations lead the system to assign high similarity scores to incorrect matches, producing tightly clustered high scores among the top candidates. The second is weak distinctiveness, which occurs when even correct matches achieve similarity scores only slightly higher than competitors due to variations in illumination, weather, or viewpoint.

We capture these distributional patterns by analyzing two key factors: the local competition among top-ranked candidates and the global separation between the best match and the remaining ones. This allows for training-free uncertainty estimation applicable to any VPR method.

Our uncertainty metrics analyze the distributional properties of top-$k$ similarity scores from any VPR method, as illustrated in Figure \ref{fig:block-diagram}. We introduce three uncertainty metrics: two individual metrics capturing complementary aspects of uncertainty, and a unified metric that integrates both to provide robust performance without requiring validation data for metric selection.

\subsection{Ratio Spread ($RS$)}
It measures how strongly alternative candidates compete with the top-ranked match by evaluating how closely their similarity scores approach the highest score. It is calculated as the average ratio of each subsequent candidate’s score (from rank 2 to $k$) to the top score, reflecting the degree of competition among the most similar matches.

\begin{equation*}
    RS = \frac{1}{k-1}\sum_{i=2}^{k} \frac{s_i}{s_1}
\end{equation*}

Where $s_i$ is the similarity score of the $i-th$ ranked place in descending order.

Higher RS values indicate that several candidates achieve similarity scores close to the top match, reflecting greater uncertainty in the retrieval and a higher likelihood of perceptual aliasing. In contrast, lower RS values suggest higher confidence, as a clear margin separates the best match from its competitors. As illustrated in Figure \ref{fig:intuition}, RS aggregates information from all top-$k$ candidates rather than relying solely on the top two, making it more robust to outliers and more effective at capturing cases where multiple ambiguous matches exist.

\subsection{Similarity Distribution ($SD$)}
\begin{figure}
    \centering
    \includegraphics[width=\linewidth]{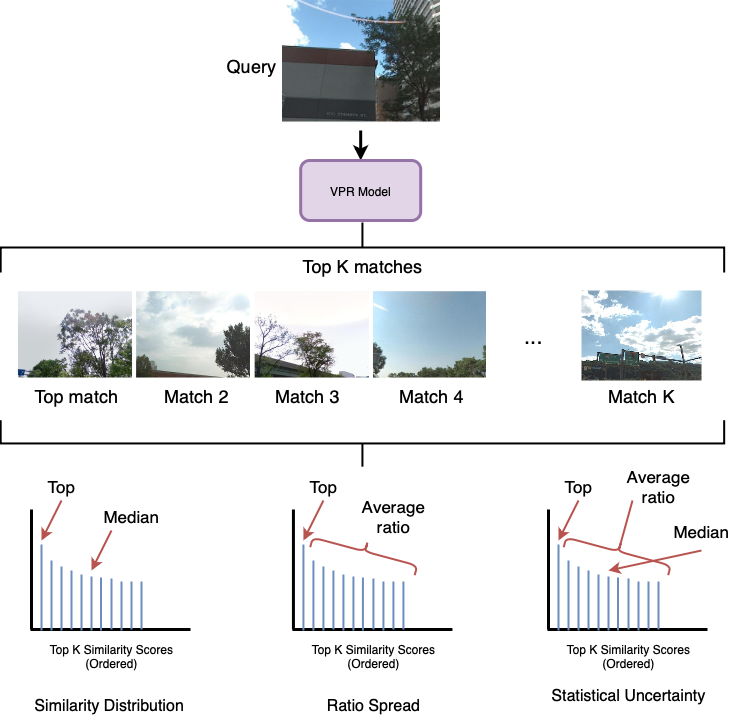}
    \caption{Visual illustration of uncertainty metrics operating on similarity score distributions. RS captures competitive ambiguity through score clustering among top-k candidates, while SD measures match distinctiveness using the distance between maximum and median scores. SU combines both metrics to provide unified uncertainty estimation.}
    \label{fig:intuition}
    \vspace{-5mm}
\end{figure}

It measures the overall dispersion of similarity scores across the top-$k$ candidates by computing the ratio of the median score to the highest score. This metric captures the global shape of the score distribution: lower SD values indicate large separations between the top match and remaining candidates, suggesting high confidence in the retrieval, while values approaching 1.0 reflect uniform score distributions where multiple candidates are equally competitive, indicating matching ambiguity. This is demonstrated in Figure \ref{fig:intuition}. Unlike methods that evaluate query-reference pairs in isolation, SD analyzes the collective distribution of top candidates, enabling more effective detection of perceptual aliasing scenarios where several visually similar but potentially incorrect matches compete with the true positive.

\begin{equation*}
    SD = \frac{s_{median}}{s_{best}}
\end{equation*}

\subsection{Statistical Uncertainty ($SU$)}
\begin{figure*}
    \centering
    \includegraphics[width=\linewidth]{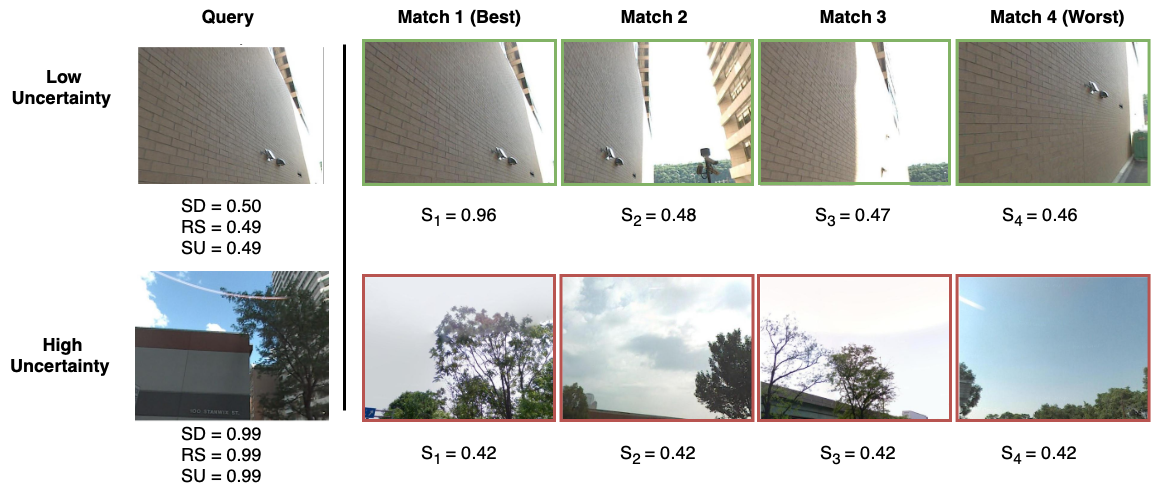}
    \caption{Visual examples of the top 4 matches for two queries using APGeM on the Pittsburgh250k dataset. Each retrieved image displays its similarity score, with the RS, SD and SU values shown below each query image. The top row demonstrates a correct retrieval with low uncertainty, while the bottom row shows incorrect matches with high uncertainty.}
    \label{fig:uncertainty_calc}
    \vspace{-5mm}
\end{figure*}

It combines two complementary distributional properties of the similarity score distribution. These metrics are combined to form the final uncertainty estimate: 
\begin{equation}
    \label{su}
    SU = \alpha \times RS + (1-\alpha) \times SD
\end{equation}
where $\alpha$ controls the relative contribution of each component. This formula captures both local competitive dynamics among top candidates and global distributional characteristics of the similarity space. Figure \ref{fig:uncertainty_calc} illustrates these uncertainty metrics applied to real retrieval examples, demonstrating how low SU scores correspond to correct matches with clear separation, while high SU scores indicate matching ambiguity, as demonstrated in Figure \ref{fig:intuition}. We validate the choice of weighting parameters through comprehensive ablation studies presented in Section \ref{ablation}.

\section{Experimental Setup}
We evaluate our uncertainty metrics across diverse VPR methods and environmental conditions using AUC-PR on benchmark datasets. All uncertainty estimation techniques are applied post-hoc to pre-computed descriptors, preserving the original performance of each VPR system at its native resolution. 

\subsection{Datasets and Methods}
We evaluate nine VPR methods: Eigenplaces \cite{eigenplaces}, Cosplace \cite{cosplace}, ConvAP \cite{convap}, MixVPR \cite{mixvpr}, AP GeM \cite{apgem}, SALAD \cite{salad}, AnyLoc \cite{anyloc}, CricaVPR \cite{cricavpr}, and NetVLAD \cite{netvlad} across six benchmark datasets: Pittsburgh 250k \cite{pitts}, St. Lucia \cite{st_lucia}, Nordland \cite{nordland}, Amstertime \cite{amstertime}, Eynsham \cite{eynsham}, and Tokyo24/7 \cite{tokyo247}. These datasets exhibit environmental diversity (seasonal changes, illumination variations, weather) and geometric challenges (viewpoint changes, perceptual aliasing, repetitive structures). Dataset details and ground-truths are provided in \cite{berton2021}.

\subsection{Evaluation Metrics and Baselines}
To ensure a fair and interpretable evaluation, we assess the uncertainty estimators independently from the retrieval backbone, thereby isolating their ability to rank predictions by confidence. We evaluate two separate but complementary aspects: 

Predictor Quality: How well does each uncertainty metric identify incorrect matches? This is measured using the Area Under the Precision-Recall Curve (AUC-PR). For this evaluation, we treat the problem as a binary classification task: "correct match" vs. "incorrect match," where the uncertainty score acts as the classifier's output. A high AUC-PR indicates the metric successfully assigns low uncertainty to correct matches and high uncertainty to incorrect ones. This directly evaluates the quality of the confidence estimate itself \cite{vpr_bench,hendrycks2018,malinin2018}.

End-to-End VPR Impact: Does using the uncertainty metric to reject predictions improve the VPR system's real-world performance? We measure this by generating precision-recall curves for the overall VPR system. By setting a threshold on the uncertainty score and rejecting all predictions deemed too uncertain, we can analyze the trade-off between precision and recall. A superior uncertainty metric will allow the VPR system to achieve higher precision at any given level of recall. Notably, this evaluation pertains exclusively to the uncertainty estimation module (Stage 3 in Figure~\ref{fig:block-diagram}), rather than the feature extraction or retrieval components of the system.

We compare against representative methods from each category:
1) Retrieval-based: L2 Distance \cite{l2} and Perceptual Aliasing (PA) \cite{pa}; 2) Spatial: Spatial Uncertainty Estimation (SUE) \cite{SUE}; 3) Learning-based: Bayesian Triplet Loss (BTL) \cite{blt}, implemented following the original paper's pseudo-code and trained on Pittsburgh250k; 4) Geometric Verification: For SIFT \cite{sift} and DELF \cite{delf}, features are extracted and then matched using a nearest-neighbor search with a traditional Lowe's ratio test for matching, while for SuperPoint \cite{superpoint}, its extracted features are matched using the SuperGlue deep-learning-based matcher \cite{superglue}. Following the matching stage, the RANSAC algorithm is employed in all three pipelines to identify the set of inlier matches. The final uncertainty value is then quantified as the negative of the total inlier count ($U_{GV} = -c_{inliers}$).

\subsection{Configuration}
Our metrics analyze similarity scores from the top 10 ($k=10$) retrieved candidates using cosine distance via Faiss \cite{faiss}. The SU score combines RS and SD with equal weighting ($\alpha = 0.5$). Experiments run on NVIDIA A100 GPUs with batch size 1 to simulate real-time robotic deployment. Results are averaged across three independent runs using standard dataset splits.

\section{Results and Discussion}
This section presents the comprehensive evaluation of our uncertainty metrics. We analyze their overall performance, computational efficiency, and ability to generalize across different VPR systems.

\subsection{Overall Performance Analysis}
Table \ref{tab:uncertainty_metrics} evaluates predictor quality—how well each uncertainty metric distinguishes correct from incorrect matches, as measured by AUC-PR. A score closer to 1.0 indicates a more effective predictor,  this measures the quality of the confidence estimates themselves.

Our three metrics consistently outperform existing retrieval-based approaches (L2, PA), learning-based methods (BTL) and spatial methods (SUE), achieving higher AUC-PR scores across nearly all VPR architectures. In particular, SU achieves the best overall results with a mean AUC-PR of 0.92, demonstrating robust uncertainty estimation across diverse visual environments and architectures. This strong intrinsic performance in identifying bad matches is the foundation for improving downstream VPR reliability.

\begin{table*}[htbp]
\centering
\begin{tabular}{l|ccccccccc|c|c}
\hline
\textbf{Uncertainty Metric} & \textbf{Eigenplaces} & \textbf{Cosplace} & \textbf{ConvAP} & \textbf{MixVPR} & \textbf{APGeM} & \textbf{SALAD} & \textbf{AnyLoc} & \textbf{Crica} & \textbf{NetVLAD} & \textbf{Mean} & \textbf{Time} \\
\hline
Baseline Model (BM) & 0.91 & 0.9 & 0.87 & 0.91 & 0.62 & 0.94 & 0.81 & 0.87 & 0.71 & 0.84 & -- \\
BM + (DUE) BTL & 0.24 & 0.25 & 0.17 & 0.17 & 0.13 & 0.33 & 0.26 & 0.13 & 0.10 & 0.20 & 0.52 \\
BM + (RUE) L2 & 0.89 & 0.90 & 0.85 & 0.90 & 0.62 & 0.94 & 0.82 & 0.87 & 0.71 & 0.83 & 0.01 \\
BM + (RUE) PA Score & 0.93 & 0.93 & 0.89 & 0.92 & 0.74 & \textbf{0.97} & 0.88 & 0.93 & 0.78 & 0.88 & 0.01 \\
BM + (RUE) RS & 0.94 & \textbf{0.95} & \textbf{0.92} & \textbf{0.94} & 0.80 & 0.96 & 0.90 & 0.94 & 0.82 & 0.91 & 0.01 \\
BM + (RUE) SD & \textbf{0.95} & 0.94 & \textbf{0.92} & \textbf{0.94} & \textbf{0.82} & 0.96 & 0.90 & 0.93 & \textbf{0.83} & 0.91 & 0.02 \\
BM + (RUE) SU & \textbf{0.95} & \textbf{0.95} & \textbf{0.92} & \textbf{0.94} & \textbf{0.82} & \textbf{0.97} & \textbf{0.91} & \textbf{0.96} & \textbf{0.83} & \textbf{0.92} & 0.03 \\
BM + SUE & 0.68 & 0.68 & 0.71 & 0.81 & 0.48 & 0.88 & 0.73 & 0.88 & 0.53 & 0.71 & 0.05 \\
\hline
BM + (GV) SIFT & 0.72 & 0.72 & 0.67 & 0.66 & 0.38 & 0.83 & 0.66 & 0.72 & 0.42 & 0.64 & 139.86 \\
BM + (GV) DELF & 0.79 & 0.78 & 0.75 & 0.85 & 0.72 & 0.93 & 0.81 & 0.87 & 0.74 & 0.80 & 78.54 \\
BM + (GV) SuperPoint & \textbf{0.95} & \textbf{0.95} & \textbf{0.98} & \textbf{0.95} & \textbf{0.93} & 0.96 & \textbf{0.94} & 0.95 & \textbf{0.94} & \textbf{0.95} & 84.11 \\
\hline
\end{tabular}

\vspace{2mm}
\caption{Table 1. A comparison of uncertainty metrics for Visual Place Recognition (VPR) methods, evaluated by their AUC-PR scores averaged across all datasets. Higher scores indicate better performance, with the best results in bold. The table includes a baseline (top row), computationally intensive Geometric Verification (GV) methods for comparison (bottom rows), and the average computation time to calculate uncertainty per query (last column).}
\label{tab:uncertainty_metrics}
\vspace{-2mm}
\end{table*}

\subsection{Cross-Method Generalization}
\begin{figure*}[!h]
\includegraphics[width=\linewidth]{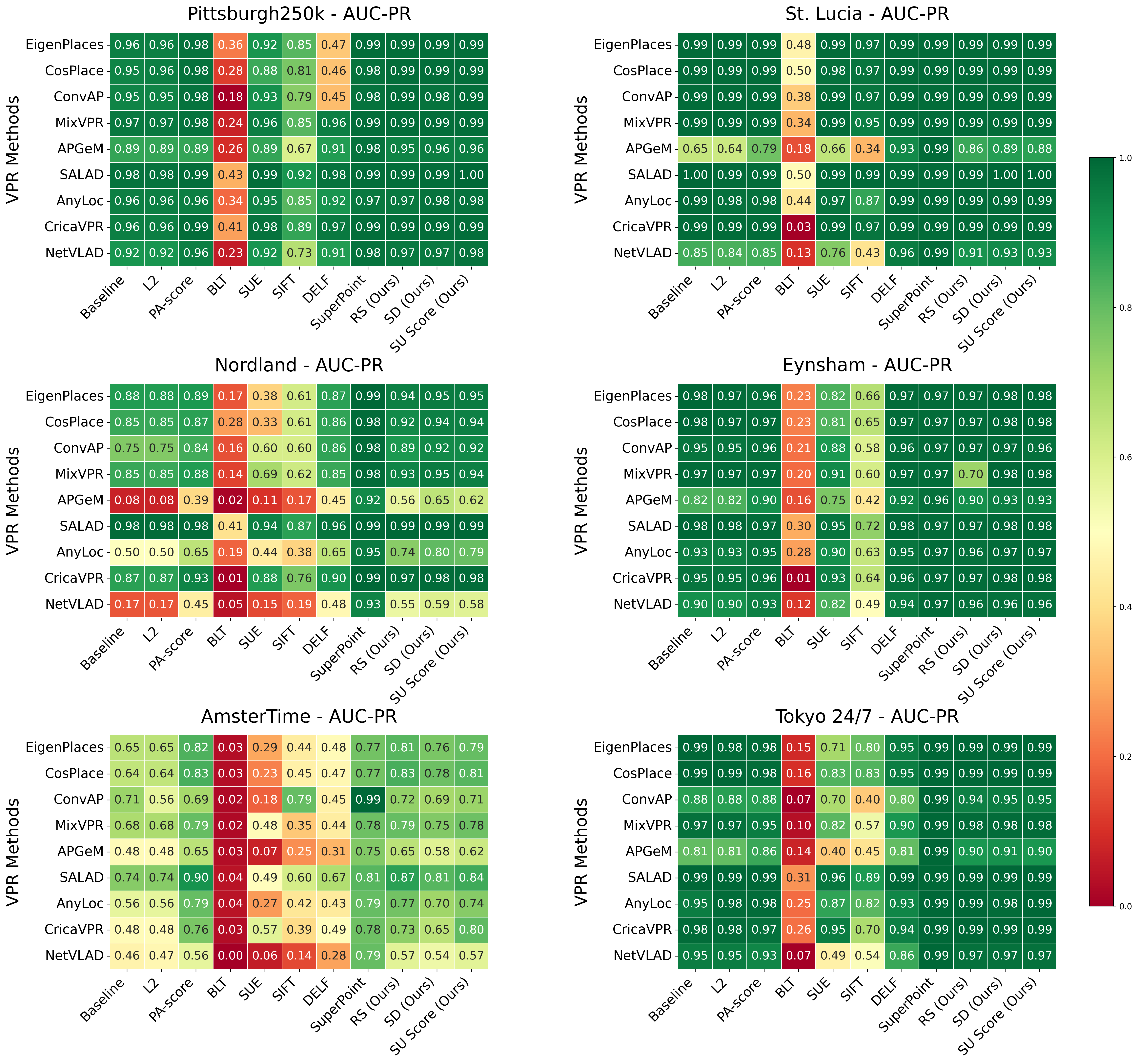}
\caption{Performance difference (AUC-PR) for each VPR and uncertainty metric combination across datasets.}
\label{fig:heatmaps}
\vspace{-5mm}
\end{figure*}

Figure \ref{fig:heatmaps} illustrates the performance gains of our metrics across different VPR architectures and datasets. Each heatmap cell shows the difference in AUC-PR relative to baseline uncertainty methods. 

Figure \ref{fig:heatmaps} visualizes the performance gains (in AUC-PR) of our metrics relative to baseline uncertainty methods across different VPR architectures and datasets. Our metrics consistently demonstrate positive gains, particularly on challenging datasets like Amstertime (viewpoint changes) and Nordland (seasonal variation). This confirms the robustness of our statistical approach across architectures, from CNN-based (NetVLAD) to modern Transformer-based models (Eigenplaces, SALAD). While geometric verification with SuperPoint achieves high accuracy, its computational cost is prohibitive for real-time use. Our metrics provide a superior balance of accuracy and speed, establishing a practical solution for VPR uncertainty estimation.

Geometric verification techniques achieve strong accuracy but are computationally expensive, making them unsuitable for real-time applications. Retrieval-based baselines (L2, PA) are efficient but less reliable across architectures. Learning-based methods (BTL, SUE) either under perform or require retraining. In contrast, our metrics achieve both superior accuracy and real-time inference speed, establishing a more practical solution for VPR uncertainty estimation.

\subsection{Computational Efficiency}
The proposed metrics are exceptionally efficient, requiring less than 0.03 ms per query (see Table \ref{tab:uncertainty_metrics}). This is orders of magnitude faster than geometric verification methods like SuperPoint (84 ms) and makes our approach suitable for real-time robotic applications where computational resources are limited. This efficiency is achieved by performing lightweight statistical analysis directly on pre-computed similarity scores, avoiding any feature extraction or matching overhead.

\subsection{Training-Free Advantage and Practical Deployment}
\begin{figure}[h]
\centering
\includegraphics[scale=0.35]{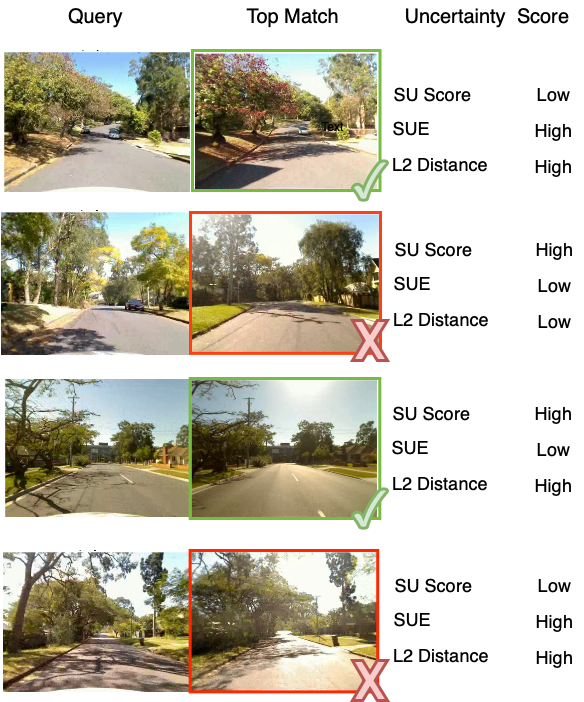}
\caption{Qualitative comparison on the Pittsburgh250k dataset. Top: correct uncertainty identification; bottom: failure cases. Query (left) and top-ranked retrieval (right) pairs shown.}
\label{fig:high_low_uncertainty}
\vspace{-5mm}
\end{figure}

The training-free nature of our metrics ensures immediate applicability to any VPR system without fine-tuning, simplifying integration across different camera setups and environments. Figure \ref{fig:high_low_uncertainty} provides qualitative examples from the Pittsburgh250k dataset, illustrating how our metrics correctly identify high uncertainty in ambiguous scenes (e.g., repetitive building facades) and low uncertainty in distinctive, correct matches.

\section{Ablation Study}\label{ablation}
This ablation study examines the effect of different $\alpha$ weighting combinations in the SU equation and evaluates the robustness of the chosen top-$k$ value.

\subsection{SU Alpha Weighting Study}
\begin{figure*}
    \centering
    \includegraphics[width=\linewidth]{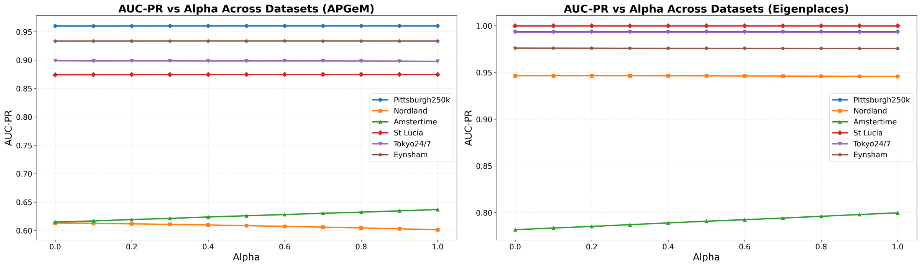}
    \caption{AUC-PR performance as a function of $\alpha$ weighting parameter (Equation \ref{su}) across multiple datasets. (Left) Results using APGeM as the VPR method. (Right) Results using EigenPlaces as the VPR method. Each line represents a different dataset, showing how the combined uncertainty metric SU performs across varying contributions of RS and SD components.}
    \label{fig:aucpr_alpha}
    \vspace{-4mm}
\end{figure*}

We evaluated the SU score's performance by varying $\alpha \in [0.0, 1.0]$ for the APGeM and EigenPlaces methods, with results shown in Figure \ref{fig:aucpr_alpha}. While performance is stable for most datasets, challenging ones like Amstertime and Nordland show preferences for either SD or RS, depending on the VPR method used. This highlights a practical challenge: the optimal individual metric can change with the deployment context. Setting $\alpha=0.5$ provides a robust default that balances both metrics, delivering consistently strong performance without requiring validation data for tuning. It prevents significant performance drops and generalizes well across diverse conditions.

\subsection{K-Value Robustness Analysis}\label{k_choice}
This ablation study examines the influence of the top-$k$ candidates used in our uncertainty metrics, evaluated on the Pittsburgh250k dataset across multiple VPR methods, with findings that generalize consistently to other datasets.

\begin{figure*}[!h]
    \centering
    \includegraphics[width=\linewidth]{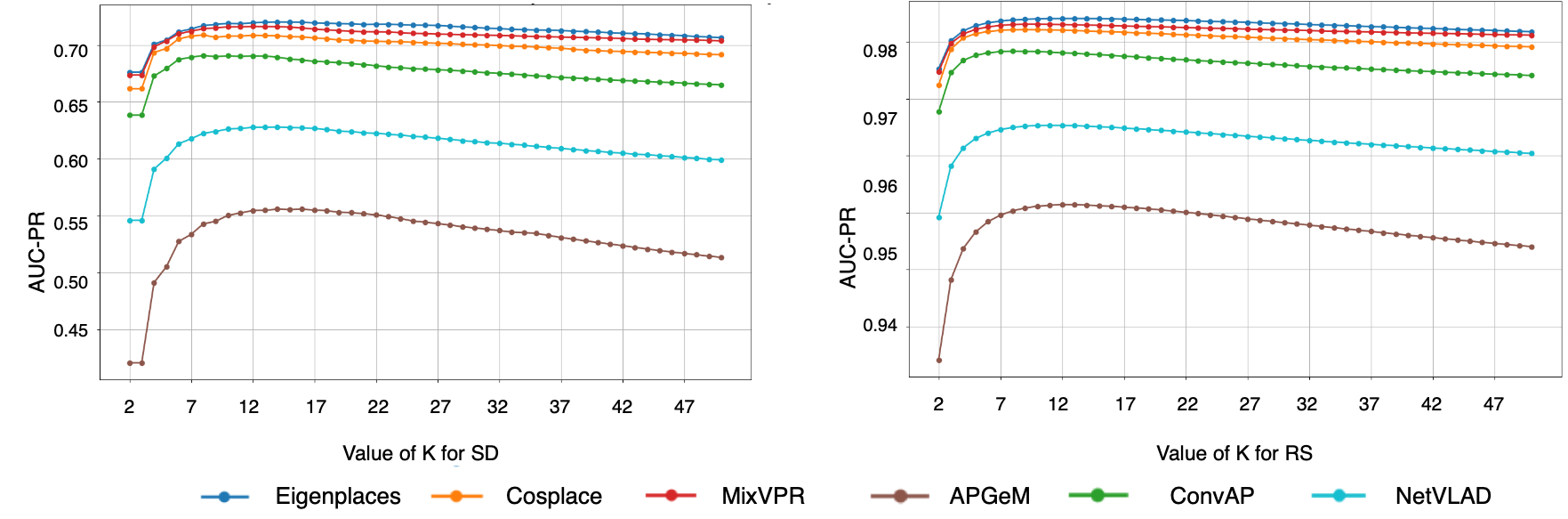}
    \caption{Comparison of $SD$ and $RS$ uncertainty metrics across $k$ values for VPR methods on Pittsburgh250k. AUC-PR results show both SD and RS provide more consistent performance across $k$ values.}
    \label{vary-k}
    \vspace{-5mm}
\end{figure*}

We analyzed the influence of the number of top candidates ($k$) on the Pittsburgh250k dataset. As shown in Figure \ref{vary-k}, both SD and RS show stable performance for $k \geq 5$, with performance stabilizing around $k=10$ and plateauing beyond that. This robustness allows for smaller k values in resource-constrained settings without compromising estimation quality and avoids the need for method-specific tuning.


\section{Conclusions and Future Work}
This paper introduced three training-free uncertainty metrics for VPR: Ratio Spread (RS), Similarity Distribution (SD), and their unified combination, Statistical Uncertainty (SU). Our key contribution is a method-agnostic approach that requires no training, architectural modifications, or validation data for metric selection, while adding negligible computational overhead ($<$0.03 ms per query).

Our comprehensive two-stage evaluation demonstrated that SU not only excels at identifying incorrect matches (0.92 mean AUC-PR) but also translates this capability into tangible improvements for the end VPR system, enhancing precision in applications like loop-closure detection. While these metrics are highly effective, their reliance on similarity scores means performance may degrade in extremely repetitive scenes where geometric cues are necessary. Future work will explore lightweight integration of geometric or semantic information to further improve robustness and investigate adaptive weighting of the RS and SD components based on score distributions.


\bibliographystyle{IEEEtran}
\bibliography{refs}

\end{document}